\documentclass[sigconf]{acmart}
\hypersetup{breaklinks=true}
\usepackage{amsmath}
\usepackage{graphicx}
\usepackage{subcaption}
\usepackage{multirow}

\AtBeginDocument{%
  }
    
\setcopyright{acmlicensed}
\copyrightyear{2025}
\acmYear{2025}
\acmDOI{XXXXXXX.XXXXXXX}
\acmConference[WEBSCI '25]{ACM Web Science Conference}{May 20-23, 2025}{New Brunswick, NJ}
\acmISBN{TBD}

\begin{document}

\title{DocNet: Semantic Structure in Inductive Bias Detection Models}


\author{Jessica Zhu} 
\email{jeszhu@umd.edu}
\orcid{0000-0001-8729-511X}
\affiliation{%
  \institution{University of Maryland}
  \city{College Park}
  \state{MD}
  \country{USA}
}

\author{Michel Cukier} 
\email{mcukier@umd.edu}
\affiliation{%
  \institution{University of Maryland}
  \city{College Park}
  \state{MD}
  \country{USA}
}

\author{Iain Cruickshank}
\email{icruicks@andrew.cmu.edu}
\affiliation{%
  \institution{Carnegie Mellon University}
  \city{Pittsburgh}
  \state{PA}
  \country{USA}
}

\renewcommand{\shortauthors}{Zhu et al.}

\begin{abstract}
News will be biased so long as people have opinions.  As social media becomes the primary entry point for news and partisan differences increase, it is increasingly important for informed citizens to be able to recognize bias. If people are aware of the biases of the news they consume, they will be able to take action to avoid polarizing echo chambers. In this paper, we explore an often overlooked aspect of bias detection in media: the semantic structure of news articles. We present DocNet, a novel, inductive, and low-resource document embedding and political bias detection model. We also demonstrate that the semantic structure of news articles from opposing political sides, as represented in document-level graph embeddings, have significant similarities. DocNet bypasses the need for pre-trained language models, reducing resource dependency while achieving comparable performance. It can be used to advance political bias detection in low-resource environments. Our code and data are made available at: https://anonymous.4open.science/r/DocNet/

\end{abstract}


  
\received{07 December 2024}

\maketitle

\section{Introduction}
The news media ecosystem has transformed dramatically in the past decade. Whereas previously only well established, professional news sources could easily disseminate information, now any outlet with an internet connection and social media presence can share their information to a broad audience \cite{RodrigoGins2023ASR}. Social media has allowed not just the proliferation of fake news, but also the exacerbation of polarizing echo chambers. Users are increasingly and unwittingly only exposed to and interact with media that already aligns with their biases \cite{Cinelli2021}. As such, to maintain a well-informed society, readers need to be aware of biased reporting so that they can also know to seek out non-partisan or diverse reporting \cite{Spinde2022HowDW}. It is imperative that resources for automatic article level bias detection be readily available. 

Bias detection remains an inexact practice, exacerbated not only by its inherently subjective nature but also by the many different forms it can manifest as \cite{RodrigoGins2023ASR}. Various datasets have been developed over the past five years that have increased research of biases at the word, sentence, article, and document level \cite{Zou2023CrossingTA, Wessel2023, Fan2019, Vallejo2023ConnectingTD, Spinde2021a, Kiesel2019, Carragher2024}. These automatic bias detection models have historically relied on curated linguistic feature-based characterizations of text \cite{Kiesel2019, Spinde2021}. More recent studies have sought to use generalizable and computationally expensive embeddings through deep learning-based methods, like contrastive learning \cite{kimjin2022}, pre-trained language models \cite{Krieger2022, Vargas2023, Spinde2022, Lei2022SentencelevelMB, Baly2020, Raza2022}, Long Short-Term Memory (LSTMs) neural networks \cite{Roy2020, Lei2022SentencelevelMB},  and Large Language Models (LLMs) \cite{Lin2024IndiVecAE}. 

These deep learning methods leverage word context. However, they do not explicitly factor in writing style structures. As such, we propose an inductive, graph-based approach to political bias detection using semantic structure. We construct independent word co-occurrence graphs for each news article and discover that their graph embeddings have comparable performance when used to detect political biases as language model embeddings. Our detection model also statistically outperforms LLM-generated bias predictions. In this paper, our contributions are threefold: 
\begin{enumerate}
    \item We provide a methodology, DocNet, for a novel, low resource, and inductive document network and political bias detection model that is not reliant on pre-trained language models.
    \item We demonstrate the value of semantic structure in bias detection within the media ecosystem.
    \item We identify high semantic structural similarity between news on either side of the political spectrum.
\end{enumerate}

\section{Background}
Bias detection is complex. Researchers have mapped out 17 different categories of media bias that have different methods and intent \cite{RodrigoGins2023ASR}. Bias detection in the media is further complicated by high text re-use in published news articles \cite{boumans} and the many different frames available to discuss an issue \cite{card2015}. A large part of widely circulated online news consists of redistributed or edited copies from news agencies. The decisions behind what to publish or how to edit an article, to include re-used text, result in instances of informational and lexical biases. Informational biases occur when news articles change the surrounding information in order to shape how an issue is framed, and thereby perceived. Lexical bias occurs when polarizing words or clauses are used to illicit specific emotions toward an issue \cite{Fan2019}. Political biases in news articles are generated through these shifts in content and structure. Identification of these rhetorical methods in a high text re-use media ecosystem requires isolating minute shifts in otherwise similar articles. 

\paragraph{News Bias Detection} Current news bias detection models use pre-trained word embeddings, like GloVe or Word2Vec, and language models, like BERT, to capture biased language in news articles. For instance, Zou et al., \cite{Zou2023CrossingTA} construct an article bias detection model using RoBerta-base and the POLITICS language model, which is a finetuned language model for ideology prediction \cite{liu2022}.  Truica et al. combine term frequency-inverse document frequency (TF-IDF), and both non-pre-trained and pre-trained language model word embeddings to represent documents for article bias detection \cite{Truica2023}.  Meanwhile, Chen et al., use the BASIL dataset \cite{Fan2019} and predict article level biases using the sentence level biases' frequency, position, and sequential order \cite{Chen2020}. They find that for the small BASIL dataset, lexical information (e.g. through bag of words), was insufficient in detecting article-level bias, while the features that characterize the internal structure of an article were more effective. These methods all hinge on either a corpus of meticulously extracted sentence level and event features, or a suitable pre-trained language model. 

\paragraph{Document Classification} Article-level bias detection can also be reframed as a document classification problem. Methods remain similar within the more generalized document classification space. Most models are highly dependent on pre-trained language models, like Longformers \cite{cohan2020} or expert generated features. Doc2Vec \cite{doc2vec} is one of the few methods that requires neither. However, transformer based approaches, where word embeddings from pre-trained language models are aggregated, have largely surpassed Doc2Vec's performance. 

Document classification researchers have also created embeddings via graph representations \cite{Wang2023GraphNN, Liu2021}. Much of graph based document classification research has been transductive, where one graph is generated for the entire corpus of documents, to include the ``test'' set. Word Nodes are connected across documents and document nodes are generated for classification \cite{Zhao2023TextGCLGC, Dai2022, Nikolentzos2020, Yao2018, Carragher2024}. Transductive models, however, are not representative of reality and are not computationally efficient as they require test documents during training. They must be retrained before being applied to new cases.

InducT-GCN is one of the few inductive document classification models discussed in literature \cite{wang2022}. It still creates one graph for the entire training corpus using document node vectors and Word Node vectors, but can update only new nodes during testing. Word-word edges are connected by their pointwise mutual information (PMI) across documents. Word-doc edges are connected using TF-IDF. They train a graph convolutional network (GCN) model to learn the word embeddings and conduct classification. These word embeddings and learned weights are used to generate predictions on batches of test documents. 

Other researchers have found that inductive document level models struggle with longer documents due to their dense graphs \cite{Piao2022}. This is apparent in the case of Induct-GCN, where while still outperforming baselines, there is distinctly worse performance on longer document datasets like 20 News Group and Ohsumed, than for shorter ones like R52 \cite{wang2022}. We address these gaps through our novel, proposed approach, DocNet. It is inductive, and therefore easily generalizable to new data. It also performs well on news articles. It achieve this through making use of only the semantic structure in text, rather than the context dependent comparisons common in the aforementioned studies.

\section{Method}
\begin{figure}[ht]
    \centering
    \includegraphics[width=\linewidth]{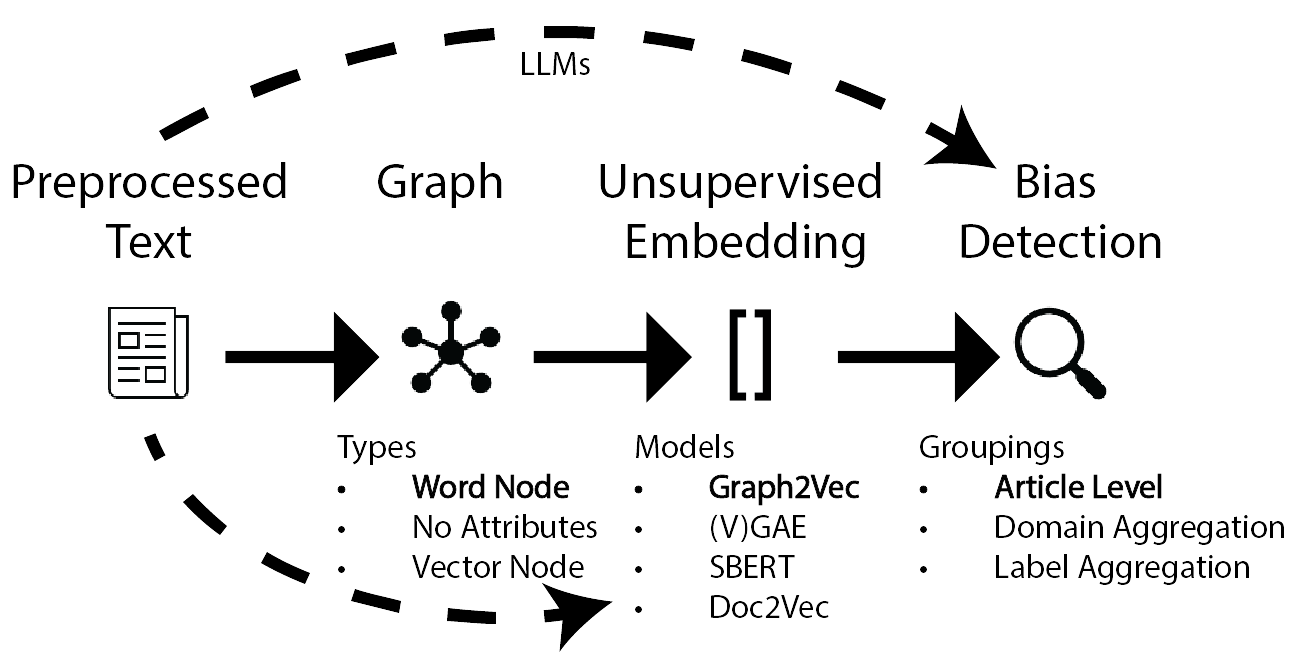}
    \caption{Pipeline of DocNet Methodology (bold) and Additional Experiments.}
    \Description{}
    \label{fig:method}
\end{figure}

In the following section, we detail our experimental methodology. The overall pipeline for our inductive bias detection methodology has four procedures, 1) Dataset collection and preprocessing, 2) Graph construction, 3) Unsupervised embedding generation and 4) Supervised model training. The DocNet pipeline and its experiments are also summarized in Figure \ref{fig:method}.

\subsection{Datasets}
Our analysis uses four datasets, three of which are centered around partisan topics: withdrawal from Afghanistan, the Oath Keepers and the military vaccine mandate, and the fourth of which is the BASIL benchmark dataset \cite{Fan2019}. For brevity we will refer to them as the AFG, OATH, VAX, and BASIL datasets, respectively. we clean and lemmatize each article.

Table \ref{tab:dataset} summarizes key characteristics of these four datasets. Tables \ref{tab:top10} and \ref{tab:bottom10} list the ten most and least frequent domains in the cleaned dataset.

\begin{table}[ht]
 
    \centering  
     \caption{Summary of datasets (Note: ``Far-Left'' labels do not exist in this data. ``Sentences'' is shortened to ``sent'')}
    \begin{tabular}{lrrrr}
        \toprule
        Dataset & AFG & OATH & VAX & BASIL \\
        \midrule
        \# of Articles & 5490 & 1965 & 1367 & 300 \\

        \# of Domains & 746 & 417 & 400 & 3 \\

        Sent/Article & 39.3 & 40.4  & 37.0   & 26.6 \\

        Words/Sent & 23.0 & 24.1 & 23.4 &  28.1 \\

        \%Left & 8.3 & 20.5 & 3.9 & 28.0 \\

        \%Left-Center & 40.8 & 48.0 & 23.6 & 00.7 \\

        \%Center & 20.2 & 23.8 & 26.1 & 46.0 \\

        \%Right-Center & 14.9 & 4.4 & 12.3 & 0.0 \\

        \%Right & 15.4 & 3.2 & 26.1 & 25.3 \\

        \% Far Right & 0.3 & 0.2 & 3.7 & 0.0 \\
        \bottomrule
    \end{tabular}
  
     \label{tab:dataset}
\end{table}

\begin{table}[ht]
    \centering  
    \caption{Top 10 Most Frequent Domains}
    \begin{tabular}{lr}
        \toprule
        Domain & Label  \\
        \midrule
        apnews.com           & left-center \\
        theepochtimes.com    & right \\
        washingtonpost.com    & left-center \\
        dailymail.co.uk      & right \\
        cnn.com              & left \\
        reuters.com           & center \\
        foxnews.com           & right \\
        msn.com               & left-center \\
        news.yahoo.com        & left-center \\
        theguardian.com       & left-center \\
        \bottomrule
    \end{tabular}
     \label{tab:top10}
\end{table}

\begin{table}[ht]
    \centering  
    \caption{10 Least Frequent Domains}
    \begin{tabular}{lr}
        \toprule
        Domain & Label  \\
        \midrule
        mb.com.ph                & left-center \\
        idahopress.com           &   left-center\\
        icrw.org                 &  left-center\\
        statesville.com          & center\\
        i24news.tv               & center\\
        humansarefree.com        & right\\
        humanevents.com          & right\\
        steadfastandloyal.com    &  right\\
        hrc.org                  & left\\
        hoover.org               & right-center\\
        \bottomrule
    \end{tabular}
     \label{tab:bottom10}
\end{table}

\paragraph{Topic Aligned Datasets}
We introduce three topic-based datasets, the AFG, Oath, and VAX news articles, and have made them publicly available. We chose these three topics for experimentation due to the significant attention, controversy, and corresponding media coverage surrounding each. Each topic's respective articles are collected by first pulling Tweets from February 1st, 2022 until December 29nd 2022. These Tweets are pulled using a keyword search of relevant terms via Twitter's Researcher API.  Table \ref{tab:collection_terms} displays the collection terms used.

\begin{table}[ht]
    \centering  
    \caption{Collection Terms for Each Dataset}
    \begin{tabular}{ll}
        \toprule
        Dataset & Collection Terms \\
        \midrule
        AFG & \{afghanistan, taliban, kabul, bagram \\
        & U.S. withdrawal, afghan\} \\
        OATH & \{oath keepers\}  \\
        VAX & \{vaccine, vax\} \\
        \bottomrule
    \end{tabular}

    \label{tab:collection_terms}
\end{table}

Per the methodology of other works \cite{cruickshank2020clustering, cruickshankcarley}, we extracted URLs from Tweets and filtered out social media site URLs to ensure only news media sites remained. We next scraped the textual content of the websites that allowed automated scraping. This methodology results in a variety of news sources and articles that are relevant to discussions on these partisan topics \cite{cruickshank2023analysis}. We align the collected domains to labels from a dataset of domains and their biases compiled by Cruickshank and Carley \cite{cruickshankcarley}. Domains are the publishing sites that news articles are scraped from. The term domain is used in lieu of news sources, as articles are often shared from distributing news agencies, like the Associated Press, to multiple sites (i.e. domains).

A weak bias label (far left, left, left-center, center, right-center, right, far right) for each article is assigned based upon its domain's bias label. Our usage of domain labels as weak labels for article bias follows the assumption that for highly contentious topics, the bias of a news article and the events that are discussed are highly correlated with the perspective of its publisher, or domain \cite{Zou2023CrossingTA}.  We further validate the reliability of the domain bias labels by comparing them with the domain bias levels published on Adfontes, as of 10 March 2023\footnote{\url{https://adfontesmedia.com/}}. Per a Pearson correlation test of news domain bias labels using an ordinal encoding, we validate that our labels are highly correlated  with Adfontes' with a correlation coefficient of .79 and p-value $<e^{-80}$. This is in spite of the three year difference in times at which these domain bias labels were collected (2020 vs 2023) and the likely drift in media bias over time. This test increases confidence that the domain bias labels we use have minimal biases themselves, as both data sources report similar labels. 

\paragraph{BASIL}
We further validate our approach with a dataset that has human annotated article and domain level biases, the BASIL dataset \cite{Fan2019}. The BASIL dataset is composed of 300 articles from 2010 to 2019 that cover 100 events, with one article each from Fox News, New York Times, and Huffington Post. These domains are considered to be right, center, and left, respectively.

\subsection{Graph Construction}
Next, in order to represent each article's semantic structure, the syntactical patterns used to create meaning, we transform each news article into its own graph.  Each graph is an undirected word (or lemma) co-occurrence network. An edge is drawn between co-occurring words in a sentence, where the weight of the edge is the positive pointwise mutual information (PPMI), Equation \ref{eq:PMI}. The edge is removed if there is a non-positive PMI in order to narrow down edges to connections with a strong association. 

\begin{equation}
    \begin{split}
    PMI = \frac{log_{2}(P_{w_1w_2}/(P_{w_1}*P_{w_2}))}{-log_{2}(P_{w_1w_2})}) \\
    P_{x} = \frac{\text{Frequency of $x$ in document}}{\text{Total Unigrams or Bigrams}} \\
    \text{$w_1$, $w_2$ are the words that construct edge $(w_1, w_2)$}
    \end{split}
    \label{eq:PMI}
\end{equation}

We construct three variations of these article graphs:
 
 \begin{enumerate}
     \item Base: No node attributes are provided.
     \item Word Nodes: The node attributes are the unique word. This is the DocNet graph.
     \item Vector Nodes: The node attribute is the word embedding vector per Spacy's en\_core\_web\_lg model\footnote{Version 3.4.1} \cite{spacy2} and concatenated with the average sentiment of the given lemma across all occurrences in the article per NLTK VADER's Sentiment Intensity Analyzer \cite{Hutto_Gilbert_2014}. 
\end{enumerate}

\subsection{Unsupervised Embeddings}
We then embed each article using unsupervised models. We experiment with 5 different embedding methods, two of which embed the graph representation of each article, two of which embed the article's cleaned text, and one hybrid approach. Except for SBERT, which is pre-trained, these embedding models are inductively trained with the same 80\%/20\% train/test splits. We do not conduct hyperparameter tuning. Instead, we seek to align parameters, like epochs and dimension size, across models for more equitable comparison, and otherwise use default package hyperparameters. No notable performance improvements were observed during initial experiments to warrant the additional computational cost of hyperparameter tuning.

\paragraph{Graph2Vec}
We train three Graph2Vec models for each graph variation using the KarateClub package \cite{karateclub}. Graph2Vec conducts Weisfeiler-Lehman (WL) hashing to generate graph level features that are then passed as the ``document'' in Doc2Vec \cite{doc2vec}. In the base graph's case, the hashing step is conducted using node degrees, whereas the designated node attribute is used otherwise. The PPMI scores are not factored into the model as Graph2Vec does not currently support edge attributes. We set a dimension size of 128, 50 epochs, and 2 Weisfeiler-Lehman iterations.

\paragraph{Graph Auto-Encoders}
We also train graph auto-encoder models (``GAE'') and variational graph auto-encoder models (``VGAE'') \cite{kipf2016} on the Vector Node graph representation of the text, both with and without the sentiment vector (``sent''). Both graph auto-encoders are constructed using three modules of graph convolution, graph normalization, and rectified linear unit (ReLU) layers in the encoder, and an inner product decoder. In order to have a comparable embedding size and epochs to the other models, we set output feature size to be 128. We use an Adam optimizer with a starting learning rate of .1 and a scheduler that reduces the learning rate when the loss plateaus after a minimum of five epochs and maximum of fifty epochs.

\paragraph{Doc2Vec}
As a comparison to the graph based approaches, we use Doc2Vec \cite{doc2vec}, using both the distributed memory (``Dm'') and the distributed bag of words (`'Dbow'') representations. We set a window size of 10, initial learning rate of .05, 50 epochs, minimum word count to be 3, negative sample of 5, and a vector size of 128.

\paragraph{Doc2Vec and Graph2Vec}
In order to measure if graph representations capture anything different from document representations, we also concatenate the separately trained Graph2Vec and Doc2Vec embeddings into a 256 dimension sized vector. We combine the two variants of Doc2Vec with the three variants of Graph2Vec for a total of six different embedding models. 

\paragraph{Sentence BERT}
As a final unsupervised embedding comparison, we use Sentence BERT (SBERT) \cite{sbert}. For each sentence in an article, we pass it through the pre-trained ``All-Distilroberta-V1'' encoder to map it to a default 768 dimensional vector space. This language model is a top performing sentencing language model of modest size. Each article is represented by the average of all of its sentence embeddings. 

\subsection{Bias Detection}
We explore the value of these embeddings in political bias detection using a naive multinomial logistic regression model on the same train/test splits as the unsupervised embeddings. Along with training a supervised model at the article level, we also experiment with domain aggregation, and bias label aggregation.  These variations allow us to explore the level of granularity of the bias information captured in the various embeddings. It also allows us to interrogate the potential for similarities between opposing viewpoints in partisan discussions. The results from our pipeline are also compared with those from LLMs and a naive baseline. 

\paragraph{Domain Aggregation}
We explore training domain level bias detection models due to the weak article bias labels used in the AFG, OATH, and VAX datasets. In the topic aligned datasets, there is a mixed number of articles between domains. As a result, we deduplicate domains  for training a news domain bias detector using various methods of aggregating article embeddings. In other words, the domain bias detectors are trained on a combined representation of each domain's article embeddings. 

We apply three methods of aggregating articles for each domain. First, we aggregate articles in a given domain by taking the arithmetic mean of the embeddings (``Mean''). We also explore the effect of normalizing the embeddings by its original topic. We subtract the average embeddings within a given topic from the embedding of each article  and take the average (``Topic Diff-Avg'', Equation \ref{avgtop}), and for another method, take the Frobenius norm (``Topic Diff-Norm'', Equation \ref{fronorm}). The latter two aggregation techniques aim to remove any topic based skew in the data. 

\begin{align}
    \begin{split}
        T_t = \frac{1}{n_t}\sum_{i=1}^{n_t}x_{ijt} \\
        \hat{x}_{ijt} = x_{ijt}-T_t \\
    \text{avgTopicDiff: } d_j = \frac{1}{n_j}\sum_{i=1}^{n_j}\sum_{t=1}^{n_t}\hat{x}_{ijt} \\
    \end{split}
    \label{avgtop}
    \\[2ex]
    \begin{split}
    \text{froNormTopicDiff: } d_j = \sqrt{\sum_{i=1}^{n_j}|\hat{x}_{ijt}|^2} \\
    \end{split}
    \label{fronorm}
\end{align}

    \noindent$d_j = \text{embedding of domain $j$}$, \\
    $x_{ijt} = \text{embedding of article $i$ from domain $j$ and topic $t$}$, \\
    $n = \text{number of articles}$, and \\
    $T_t = \text{average topic embedding of topic $t$}$

For the BASIL dataset, our domain level models predict the domain bias labels for each article without aggregating embeddings since there are only three different news domains. This increases the risk of the model capturing spurious characteristics, but in reviewing our cleaning procedure, we believe the risk of domain specific text cues being present is sufficiently mitigated. 

\paragraph{Label Aggregation} 
We train three separate inductive models using varied bias label groupings 1) ``Full'': the original bias categories, 2) ``LCR'': Re-classifying mildly or extremely partisan articles into right or left, for three nominally encoded bias categories (right, center, left) and 3) ``Binary'': Re-classifying into biased or unbiased, where left/right-center are both considered to be unbiased.

After initial predictions are generated, we regroup the ``Full'' category predictions into``LCR'' and ``Binary''  to more accurately compare the accuracy of models with different numbers of categories.

\paragraph{Large Language Models}
As a final comparison, we conduct document level bias detection using pre-trained large language models. We experiment with both OpenAI's GPT Turbo 3.5, which is optimized for chat completion but has also demonstrated strong performance for general purpose tasks \cite{gpt35} and Meta's Llama 3 8B Instruct Model, which is tuned for instruction tasks \cite{llama3modelcard}. We pass the following zero-shot prompt template to generate bias predictions: ``Please classify the following news article by its political bias. Please only classify the article as \{bias labels\}, and return no other text.
title: \{title\}
article: \{article\}
bias: ''.  We do not conduct domain level predictions using the large language models.

\paragraph{Naive Baseline}
We use a Zero Rule baseline for a naive model standard, where the test set predictions are all labeled as the majority class in the training set.

\section{Results and Discussion}
We ran 1284 experiments to train multiple unsupervised embeddings and supervised learning models over the various configurations and each of the four datasets. We also train a topic agnostic dataset that merges the AFG, OATH, and VAX datasets. The experiments are trained on 1 node of an NVIDIA DGX A100 system. Each graph autoencoder embedding model takes at most 2 GPU hours to train one model, while each Graph2Vec embedding model takes at most 3.5 CPU minutes. In order to control for different configurations, we formally evaluate the difference between configurations and embeddings using a one-sided Wilcoxan signed rank test on the accuracy and macro F-1 scores. As each pairwise configuration uses the same test sets but are trained independently, the necessary assumptions for a valid hypothesis test are satisfied.  This test allows us to determine if there is a paired difference between two methods when all other configurations are held constant.

Through this hypothesis test, We confirm that the DocNet embeddings outperform the naive baseline, with p-values close to 0. They have an average .21 increase, or a 120\% improvement in macro F-1 scores over the baseline. In addition, DocNet outperforms all the embedding models, except SBERT. However, in 30\% of the configurations, it also outperformed the SBERT embeddings. For the sake of brevity, we display the average and standard deviation of scores across configurations in Table \ref{tab:results}. 

\begin{table}[ht]
\centering 
\caption{Accuracy and Macro F-1 by Embedding Configurations. Avg (std)}
\begin{tabular}{lrr}
\toprule
Model & Accuracy &  Macro F-1 \\ 
\midrule
Naive Baseline & 0.522 (.159) & 0.246 (.133) \\
\multicolumn{3}{c}{LLMs} \\
GPT 3.5 Turbo & 0.417 (.164)  & 0.349 (.177) \\
\textit{Llama 3-8B} & \textit{0.454 (.225)} & \textit{0.378 (.204)}\\
\multicolumn{3}{c}{Text Embeddings} \\
\textbf{SBERT} & \textbf{0.607 (.154)} & \textbf{0.542 (.162)} \\
Doc2Vec: Dm & 0.539 (.170) & 0.472 (.170)\\
Doc2Vec: Dbow & 0.574 (.170) & 0.486 (.170) \\
\multicolumn{3}{c} {Graph Embeddings} \\
VGAE & 0.470 (.155)& 0.362 (.135)\\
GAE & 0.538 (.146) & 0.425 (.138)\\
VGAE: w/ sent & 0.475 (.143) & 0.358 (.128) \\
GAE: w/sent & 0.524 (.146) & 0.391 (.135)\\
Graph2Vec: Base & 0.498(.169) & 0.341 (.137) \\
Graph2Vec: Vector Node & 0.589 (.154) & 0.501 (.161) \\
\textit{Graph2Vec: Word Node} & {0.592 (.151)} & {0.507 (.157)}\\
\multicolumn{3}{c}{Hybrid: Doc2Vec+Graph2Vec}\\
Dbow + Base & 0.559 (.169) & 0.457 (.170) \\
Dm + Base & 0.544 (.178) & 0.467 (.184)\\
Dm + Word Node & 0.583 (.164)  & 0.501 (.167)\\
Dbow + Vector Node & 0.585 (.163) & 0.501 (.165)\\
Dm + Vector Node & 0.572 (.157)  & 0.509 (.160)\\
\textit{Dm + Word Node} & \textit{0.570 (.164)}  & \textit{0.511 (.167)}\\
\bottomrule
\end{tabular}
\label{tab:results}
\end{table}

\paragraph{Label Aggregation}
As expected, there is a large drop in performance as the number of classes increase, with the models trained for fine grained bias classification contributing to the lower overall average scores. The results from models grouped by various label bins are in Table \ref{tab:labelresults}. We also find that after regrouping the predictions from models trained using ``Full'' labels, there is not a statistically significant difference in macro F-1 in comparison to models trained using the ``LCR'' or ``Binary'' labels, with p-values of .91 and .36, respectively. This demonstrates that while the ``Full'' multiclass models have low scores, they are able to extract characteristics of bias in news just as effectively as the models trained on fewer categories. A one-vs rest model training approach would be needed to validate if more granular bias detection is possible with these embeddings.

\begin{table}[htbp]
    \centering 
    \caption{Accuracy and Macro F-1 by Label Aggregation Methods versus the Naive Baseline. Avg (std) of all experiments.}
    \begin{tabular}{lrr}
    \toprule
    Label Aggregation \\
    & Accuracy &  Macro F-1  \\
    \midrule
    Full \\
    \newline Models & 0.404 (.102) & 0.294 (.111) \\
    \newline Baseline & 0.380 (.075) & 0.105 (.038) \\
    LCR\\
    \newline Models & 0.518 (.102) & 0.465 (.102) \\
    \newline Baseline & 0.470 (.067) & 0.2167 (.036)\\
    Binary \\
    \newline \textbf{Models} & \textbf{0.722 (.099)} & \textbf{0.608 (.118)} \\
    \newline Baseline & 0.717 (.069) & 0.417 (.024)\\
    \bottomrule
    \end{tabular}
    \label{tab:labelresults}
\end{table}

\paragraph{Domain Aggregation}
The results for the different domain aggregation comparisons are listed in Table \ref{tab:configresults}. For the BASIL dataset, the embeddings were more accurately classified into the domain level labels, though the article level predictions were still significantly better than the naive baseline, with a p-value near 0 for all metrics. The method of aggregating article embeddings also makes a significant difference on classification performance, such that domain models using Topic Didd-Norm performs the worst. The mean of article embeddings across samples of a domain has the best performance in domain level bias detection. More research is needed to identify if topic information is not significant in domain level bias detection or if a better way of extracting topic level information is warranted.

\begin{table}[htbp]
    \caption{Accuracy and Macro F-1 by Domain Aggregation Method and Dataset Subsets. Avg (std)}
    \centering 
    \begin{tabular}{lrr}
    \toprule
    & Accuracy &  Macro F-1 \\
    Domain Aggregation &  &  \\
    \midrule
    No Aggregation & 0.565 (.165) & 0.453 (.167)\\
    \textbf{Mean} & \textbf{.563 (.168)} & \textbf{0.490 (.175)} \\
    Topic Diff-Norm & 0.500 (.159) & 0.397 (.153)\\
    Topic Diff-Avg & 0.556 (.163) & 0.483 (.167)\\
    BASIL: Article Level & 0.372 (.111) & 0.332(.112) \\
    BASIL: Domain Level & 0.588 (.190) & 0.544(.196)\\
    \midrule
    Dataset Specific Models &  &  \\
    \midrule
    BASIL & 0.480 (.189) & 0.438 (.191) \\
    Full & 0.520 (.173) & 0.445 (.174) \\
    AFG & 0.549 (.157) & 0.447 (.143) \\
    VAX & 0.510 (.148) & 0.451 (.166) \\
    \textbf{OATH} & \textbf{0.560 (.161)} & \textbf{0.461 (.160)} \\
    \bottomrule
    \end{tabular}
    \label{tab:configresults}
\end{table}

\subsection{Embedding Model}
The average mean and standard deviation of scores given different embedding models are in Table \ref{tab:results}. Through the paired rank test on the full set of experiments, the GPT, Llama, graph autoencoders, and Graph2Vec with no features (i.e. the base graph) perform the worst, with no statistically significant performance difference between them. The unsupervised Graph2Vec based embeddings with node attributes are able to capture characteristics of biased text. These embeddings have performance similar to, and with the Word Node graphs, 30\% of the time, better than SBERT.  However, on average SBERT still outperforms all other embeddings. Graph2Vec models with node attributes (both with Word Nodes and Vector Nodes) outperform all Doc2Vec embedding models with p-values less than .01. When Graph2Vec embeddings with node features are combined with Doc2Vec, a statistically significant improvement (p-values less than  .01) from the original Doc2Vec based model occurs. For instance, Graph2Vec with Word Nodes concatenated with Doc2Vec with distributed memory consistently outperforms embeddings using just Doc2Vec, as observed in their average scores in Table \ref{tab:results}. We do not find a statistical difference between embedding performance based on the type of node attribute used in Graph2Vec, Word Nodes or Vector Nodes. 

Although our graph based embeddings do not consistently outperform SBERT, these results demonstrate the value of including semantic structure in bias classification. Unlike with SBERT, language models and word embeddings are unnecessary in our DocNet methodology. As demonstrated by the performance of the Word Node based embeddings, so long as there is a unique node identifier that indirectly relates unique graphs, the article graph embeddings are comparable to SBERT. This indirect connection between news articles is necessary and sufficient to have performance that is comparable to pre-trained language models.  Without factoring in edge attributes, pre-trained word embeddings, or word frequency, our document networks are able to successfully embed information for bias detection.  

In contrast, even though our computationally expensive graph autoencoder models pass both edge and node attributes in generating document embeddings, they under perform in comparison to using Graph2Vec.  A simple approximate graph isomorphism comparison through WL-hashing captures enough news article information for good bias predictions. The effectiveness of the Word Node graphs validates the value of characterizing semantic structure using document networks.

\subsection{DocNet Characteristics Case Study}
In the following section, we investigate if DocNet's Word Node graph embeddings align with standard graph features. 

\begin{figure}[ht]
    \centering
    \includegraphics[width=\linewidth]{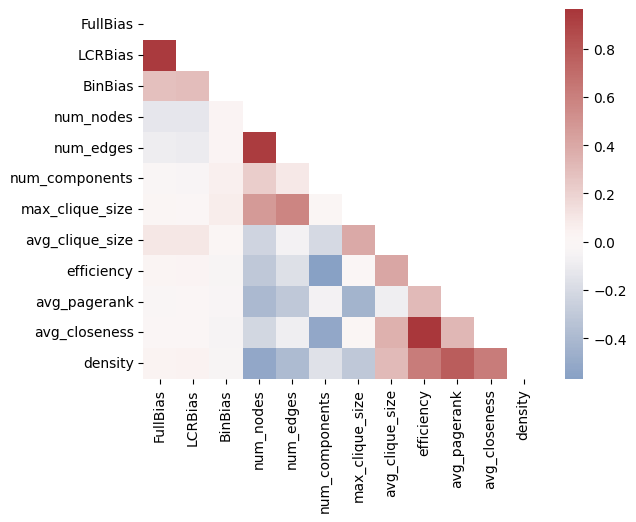}
    \caption{Correlation Plot of Graph Metrics and Bias from Topic Aligned Dataset}
    \Description{}
    \label{fig:corr_aci}
\end{figure}

\begin{figure}[ht]
    \centering
    \includegraphics[width=\linewidth]{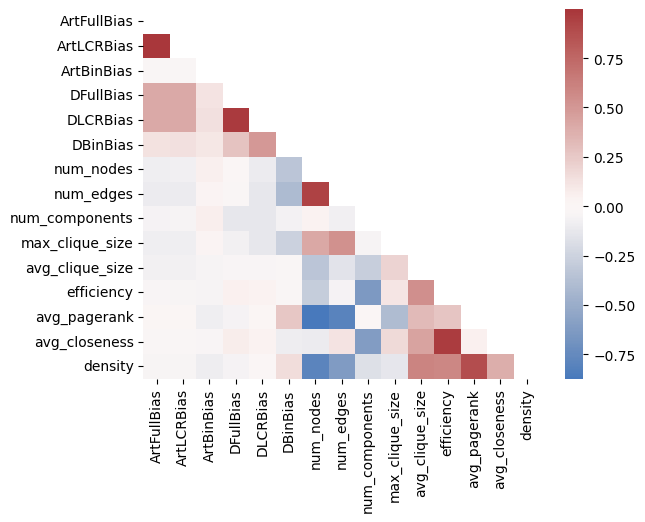}
    \caption{Correlation Plot of Graph Metrics and Bias from BASIL (Art\* is article bias, D\* is domain bias)}
    \Description{}
    \label{fig:corr_basil}
\end{figure}

We obtain descriptive graph metrics for every article and calculate their Pearson correlation with an ordinal encoding of the bias labels as shown in Figures \ref{fig:corr_aci} and \ref{fig:corr_basil}. There is no strong correlation between any graph metrics and bias labels. We then train a logistic regression model using these metrics to predict binary bias labels. These models have an average .21 lower Macro F-1 on the topic aligned datasets and a .12 lower Macro F-1 on the domain level BASIL predictions compared to the corresponding Graph2Vec with Word Node models. This analysis demonstrates that while the semantic structure of articles holds bias information, our proposed graph embeddings provide additional information over curated descriptive graph metrics.

\begin{figure}[htb]
    \centering
    \begin{subfigure}[l]{.45\columnwidth}
        \centering
        \includegraphics[width =\linewidth]{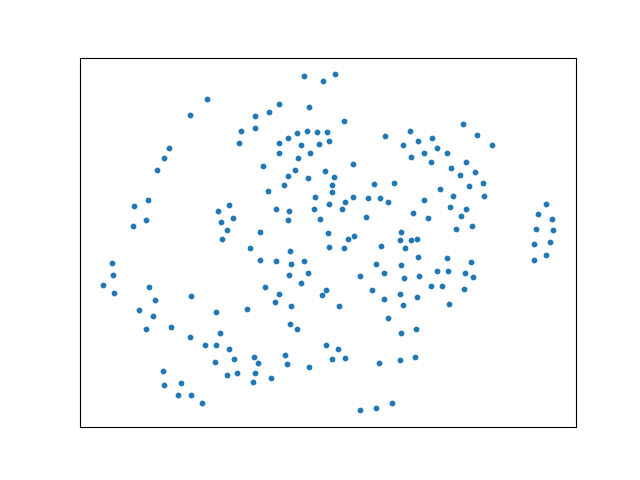}
        \subcaption{Correct-Unbiased}
    \end{subfigure}
    \begin{subfigure}[l]{.45\columnwidth}
      \centering
         \includegraphics[width=\linewidth]{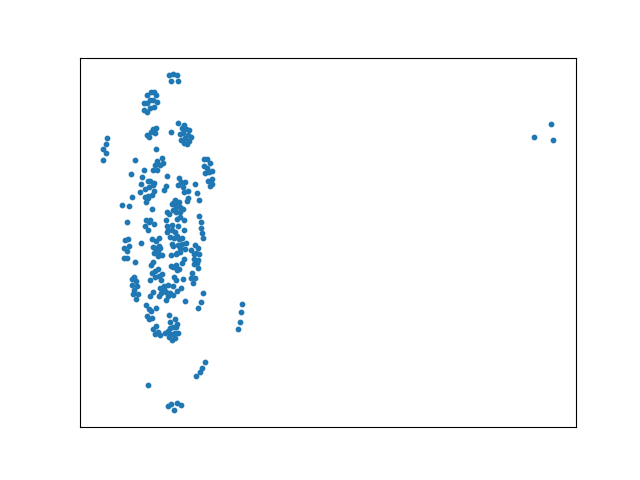}
        \subcaption{Correct-Biased}
    \end{subfigure}
    \begin{subfigure}[l]{.45\columnwidth}
       \centering
         \includegraphics[width=\linewidth]{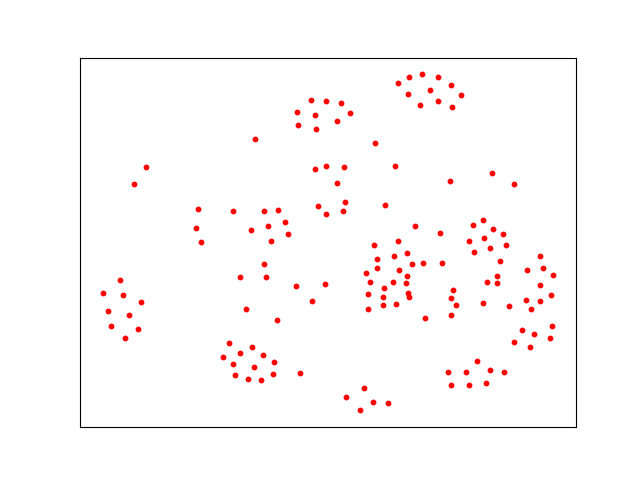} 
        \subcaption{Incorrect-True Biased}
    \end{subfigure}  
    \begin{subfigure}[l]{.45\columnwidth}
        \centering
         \includegraphics[width=\linewidth]{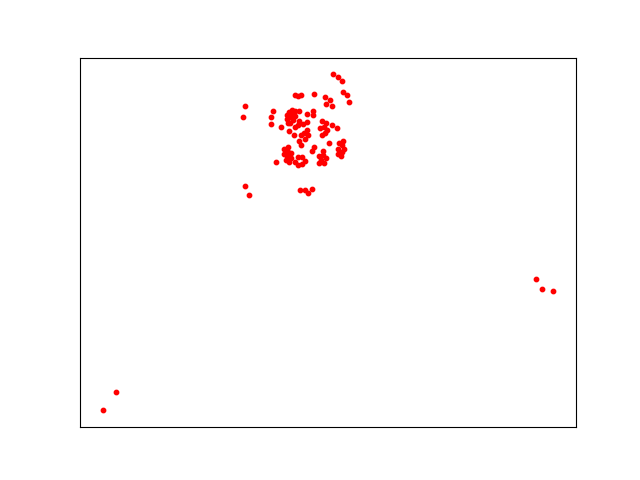}
        \subcaption{Incorrect-True Unbiased}
    \end{subfigure}
    \caption{Graphs of articles with high probability predictions from BASIL. Predictions from Graph2Vec: Word Nodes at the domain level using binary labels (Macro F-1 = .82). Source domains: a) NYTimes, b) Fox, c) Huffington Post d) NYTimes }
    \Description{}
    \label{fig:graph_basil}
\end{figure}

\begin{figure}[htb]
    \centering
    \begin{subfigure}[l]{.45\columnwidth}
        \centering
        \includegraphics[width =\linewidth]{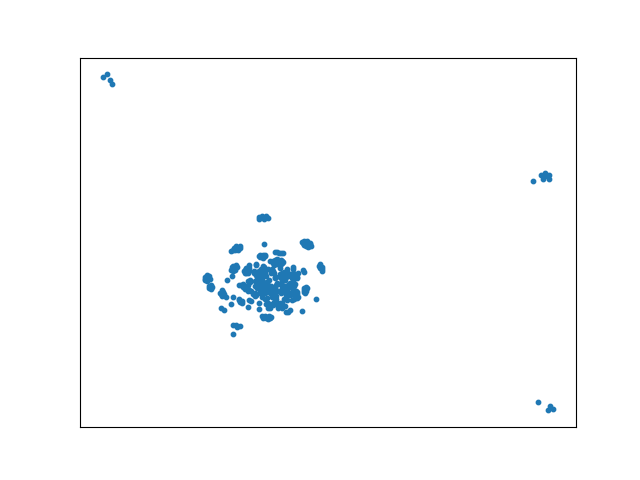}
        \subcaption{Correct-Unbiased}
    \end{subfigure}
    \begin{subfigure}[l]{.45\columnwidth}
      \centering
         \includegraphics[width=\linewidth]{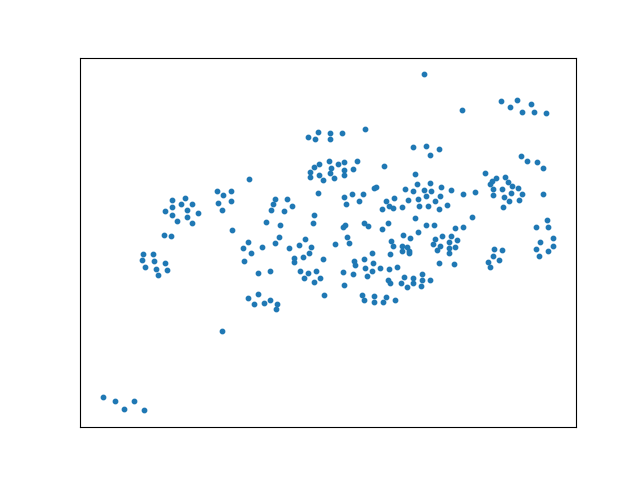}
        \subcaption{Correct-Biased}
    \end{subfigure}
    \begin{subfigure}[l]{.45\columnwidth}
        \centering
         \includegraphics[width=\linewidth]{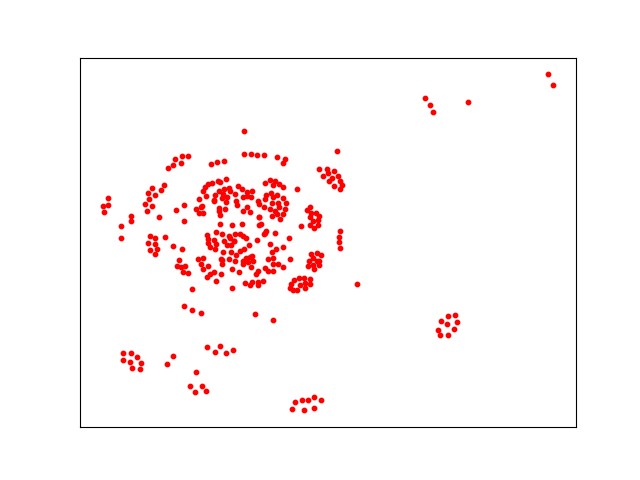}
        \subcaption{Incorrect-True Biased}
    \end{subfigure} 
    \begin{subfigure}[l]{.45\columnwidth}
       \centering
         \includegraphics[width=\linewidth]{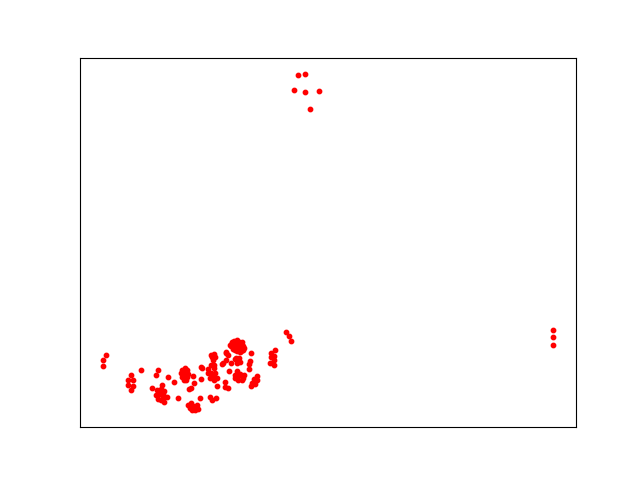} 
        \subcaption{Incorrect-True Unbiased}
    \end{subfigure} 
    \caption{Graphs of articles with high probability predictions from the VAX dataset. Predictions from Graph2Vec: Word Nodes using binary labels (Macro F-1 = .75). Source domains are: a) ABC, b) AFN, c) Toronto Sun d) MSN }
    \Description{}
    \label{fig:graph_vax}
\end{figure}

\begin{figure}[htb]
    \centering
    \begin{subfigure}[l]{.45\columnwidth}
        \centering
        \includegraphics[width =\linewidth]{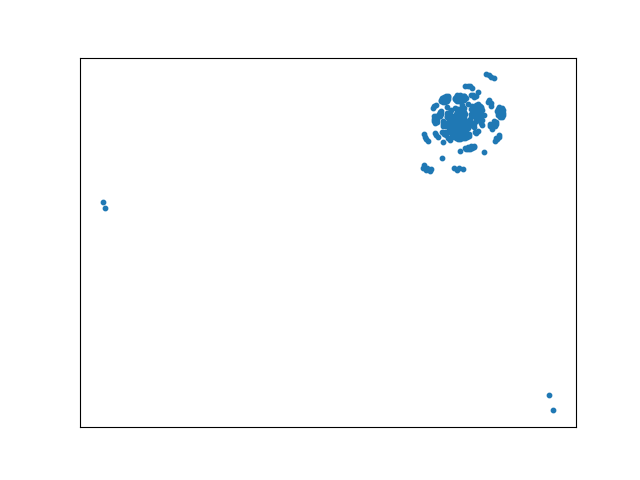}
        \subcaption{Correct-Unbiased}
    \end{subfigure}
    \begin{subfigure}[l]{.45\columnwidth}
      \centering
         \includegraphics[width=\linewidth]{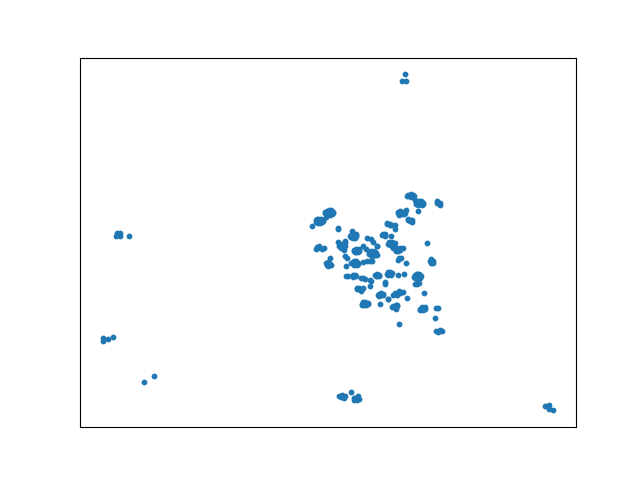}
        \subcaption{Correct-Biased}
    \end{subfigure}
    \begin{subfigure}[l]{.45\columnwidth}
        \centering
         \includegraphics[width=\linewidth]{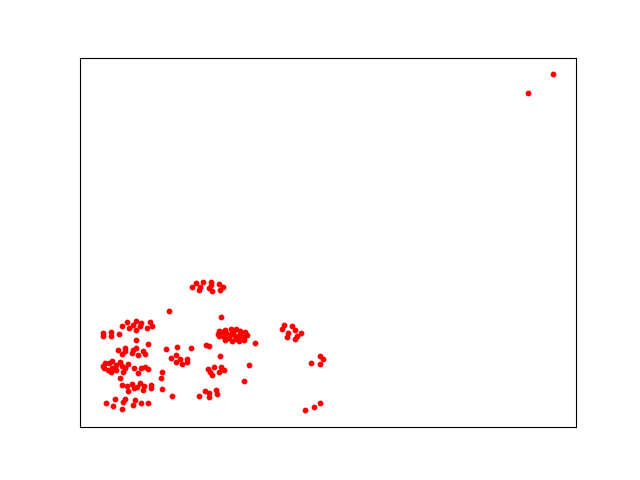}
        \subcaption{Incorrect-True Biased}
    \end{subfigure} 
    \begin{subfigure}[l]{.45\columnwidth}
       \centering
         \includegraphics[width=\linewidth]{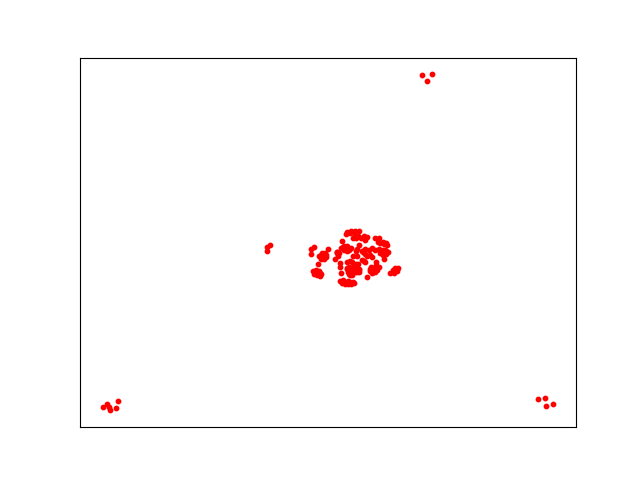} 
        \subcaption{Incorrect-True Unbiased}
    \end{subfigure} 
    \caption{Graphs of articles with high probability predictions from the OATH dataset. Predictions from Doc2Vec: Dbow + Graph2Vec: Word Nodes using binary labels (Macro F-1 = .76). Source domains are: a) MSN, b) Raw Story, c) Salon d) MSN }
    \Description{}
    \label{fig:graph_oath}
\end{figure}

We also plot graphs from top performing binary Node Word based embedding models to manually inspect for any consistent characteristics or discrepancies in the data labels. We highlight graphs from the domain label BASIL model, and no domain aggregation VAX, and OATH datasets in Figures \ref{fig:graph_basil}, \ref{fig:graph_vax} and \ref{fig:graph_oath}, respectively. These are the article graphs with the highest probability incorrect and correct predictions for both bias labels. They are graphed using a spring force-directed algorithm \cite{networkx}. 

We observe that in the graphs of BASIL articles, Figure \ref{fig:graph_basil}, the graphs for the articles that were predicted to be unbiased (a and c) have visually sparse layouts. Meanwhile, articles predicted to be biased (b and d) appear to be denser. Upon returning to the original data, we  find that the news article from Figure \ref{fig:graph_basil}c is actually labeled at the article level to be unbiased. The original article label for Figure \ref{fig:graph_basil}d also is unbiased. However Figure \ref{fig:graph_basil}d's article contains phrases like ``Donald J. Trump leads a familiar refrain'' and ``Mr. Trump, who has been criticized for lacing his campaign speeches with profanity''. These phrases may be considered to be inflammatory given its usage of words like ``refrain'' and ``lacing'' \cite{nytimesbasil}. This example demonstrates the continued difficulty in developing reliable article bias labels for an inherently subjective application. As the BASIL dataset only has three domains, these visual patterns may be from a spurious correlation.  

To interrogate these patterns further, we look at the Node Word graphs of the articles using the article level models of the VAX and OATH datasets, as shown in Figures \ref{fig:graph_vax} and  \ref{fig:graph_oath}. As is apparent in these figures, no consistent visual pattern exists between the articles with biased versus unbiased predictions. This reinforces our findings from the graph metric analysis that a graph embedding approach is needed to capture the complex semantic structure characteristics of news articles.

 Since the topic aligned datasets have significantly greater variety in domains in comparison to the BASIL dataset, the possibility the bias detection model learned spurious domain specific characteristics is reduced. In manual validation of the weak labels from Figures \ref{fig:graph_vax} and  \ref{fig:graph_oath}'s graphs, we find clear usage of informational bias. The article from Figure \ref{fig:graph_oath}b  ends with the sentence, ``Those words are directional instructions on Claymore anti-personal mines''\cite{rawstory}, which invokes a sense of incredulity of the actions of the Oath Keepers. Similarly, the AFN article from Figure \ref{fig:graph_vax}b opens with the title, ``Military docs drop vaccine bombshell but `fact-check' has an explanation'' \cite{afnvax}. The usage of ``bombshell' and quotation marks around ``fact-check'' are examples of lexical bias. Both articles align with their domain bias labels. On the other hand, the incorrectly classified OATH MSN article (Figure \ref{fig:graph_oath}d) is actually originally sourced from CNN. It contains artifacts like ``screaming match'', ``The two continued to cut each other off...'', and ``chastised''. Although its partisan lean is not immediately clear without greater context, these informational and lexical rhetorical decisions lead to the assessment that this article is biased. If the article had been correctly attributed to CNN in our dataset, it would have been aggregated as biased, rather than the unbiased, or more specifically left-center, label given to MSN \cite{cnnoath}. This inspection demonstrates that while our labeling assumptions falters on borderline biased articles, the generalizability of domain labels to weak article labels for contentious topics largely holds. Our bias detection models may even be more astute that these weak labels.

\section{Conclusion and Future Work}
Through DocNet, we demonstrate that encoding semantic structure in news articles through word co-occurrence networks can be used for domain and article level bias detection. We use a logistic regression to learn a separation of the unsupervised embedding space. Using DocNet, we also show that biased texts have similar context agnostic semantic structures. These similarities cross article topics and partisan stances. Through the inclusion of semantic structure in the form of word-co-occurrence networks, our proposed methodology has performance comparable to, and occasionally better than embeddings from transformer based language models. 

DocNet successfully uses Graph2Vec with Node Words and a multinomial logistic regression to detect bias. DocNet does not require pre-trained language models to generate word embeddings, nor does it require computationally expensive deep learning techniques. The most computationally time consuming portion of DocNet is the initial graph construction. However, this can be parallelized across a corpus of documents for time savings. It is an inductive approach, allowing for inexpensive and generalizable predictions on new datasets.

Through the varied groupings of bias labels, we find that our approach struggles with fine-grained labels. This is an observation also observed by other recent works \cite{Carragher2024}. Further research using one-vs-rest classifiers is needed to confirm if semantic structure embeddings can capture fine-grained differences. On the other hand, even with a naive supervised model, semantic structure embeddings can sufficiently characterize coarser bias labels. This also means that right-leaning news articles and left-leaning news articles have significantly more in common than either side of the political spectrum may believe. More research is needed to pinpoint exactly what these structural characteristics or writing style biases are. Regardless, we can conclude that the characteristics of partisan rhetoric are more than context-dependent. 

In future work, it would also be valuable to research alternative methods to our graph autoencoder and hybrid approaches that fuse word embeddings with semantic structure. Alternative deep learning graph models, like graph attention or diffusion based approaches \cite{Nikolentzos2020, Liu2021} could be explored for improved coupling of context and structural information. New graph hashing algorithms that efficiently encode edge attributes could also be developed and used in Graph2Vec. Future developments in inductive graph embedding approaches that efficiently and effectively represent dense networks and attributes would likely improve bias detection efforts. In the interim, our DocNet methodology provides a pathway for low-resourced bias detection in news, when computational resources, rigorously labeled datasets, or language models are not available. As an alternative to large language models, these findings may improve accessibility to underrepresented communities. It helps ensure informed news consumption as people recognize the biases in their online media ecosystems.

\section{Limitations}
As previously mentioned, this project relies on contentious topics and does not look into bias detection of issues with less polarization. In addition, since the data was collected based on articles shared over Tweets, there may be a skew in our dataset to highlight more inflammatory media with characteristics that draw the attention of Twitter users. However, given the size of the dataset, the variation in domains present, and the consistency in trends with the manually labeled BASIL dataset, we believe our data to be sufficiently generalizable. Our data should only be used for research purposes. An incorrect bias detection model may lead to misplaced mistrust or over confidence in media. Additionally, this project only analyzes articles in the English language. While our DocNet methodology is not tied to a language, future work is needed to validate its applicability to news articles from other languages. 
\balance
\bibliographystyle{ACM-Reference-Format}
\bibliography{main}


\section*{Ethics Statement}
All articles collected for this study were done so under the provisions of Section 107 of the U.S. Copyright Act and ensured that our collection action fell under the fair use category. The Tweets were collected in accordance with Twitter's terms of service at the time of collection. The code and news articles are made available at \url{https://anonymous.4open.science/r/DocNet/} under the MIT and CC0 1.0 Universal licenses, respectively.

\end{document}